\documentclass[runningheads]{llncs}
\usepackage{graphicx}

\usepackage{subcaption}

\usepackage{times}
\usepackage{latexsym}

\usepackage[T1]{fontenc}

\usepackage[utf8]{inputenc}
\usepackage{microtype}
\usepackage{amsfonts}

\usepackage{longtable}
\usepackage{xtab,booktabs}

\usepackage{tcolorbox}
\usepackage{xargs}  
\usepackage{makecell}
\usepackage{graphicx}
\usepackage{booktabs}
\usepackage{enumitem}
\usepackage{xspace}
\usepackage{color,soul}
\usepackage{amsmath}
\usepackage{array, multirow}
\usepackage{algorithm}
\usepackage{algorithmic}
\usepackage{multirow}
\usepackage{cleveref}
\usepackage{moreverb}

\usepackage{tasks}

\author{Lingbo Mo\inst{1}\thanks{~~Work done during an internship at Amazon.} \and
Besnik Fetahu\inst{2} \and
Oleg Rokhlenko\inst{2} \and
Shervin Malmasi\inst{2}}
\authorrunning{Mo et al.}

\institute{The Ohio State University, Ohio, USA \\
\email{mo.169@buckeyemail.osu.edu} 
\and
Amazon.com, Inc. Seattle, WA, USA \\
\email{\{besnikf,olegro,malmasi\}@amazon.com} 
}

\newcommand{\revision}[1]{\textcolor{black}{#1}}

\newcommand{\examplenew}[1]{``\emph{#1}''}
\newcommand{\pqar}{\textsc{PAR}\xspace}
\newcommand{\pqa}{\textsc{PQA}\xspace}
\newcommand{\reddit}{\textsc{Reddit}\xspace}
\newcommand{\semeval}{\textsc{SemEval}\xspace}
\newcommand{\smf}{\textsc{SMF}\xspace}
\newcommand{\smfs}{\textsc{SMF-Style}\xspace}
\newcommand{\mf}{\textsc{MF}\xspace}
\newcommand{\cbs}{\textsc{CBS}\xspace}
\newcommand{\tb}{\textsc{T5}\xspace}
\newcommand{\cold}{\textsc{COLD}\xspace}
\newcommand{\gpt}{\textsc{GPT2}\xspace}

\begin{document}

\title{Controllable Decontextualization of Yes/No Question and Answers into Factual Statements}

\titlerunning{Controllable Rewriting of Yes/No Question and Answers.}

\maketitle

\begin{abstract}
Yes/No or \emph{polar} questions represent one of the main linguistic question categories. They consist of a main interrogative clause, for which the answer is binary (assertion or negation).
Polar questions and answers (\pqa) represent a valuable knowledge resource present in many community and other curated QA sources, such as forums or e-commerce applications.
Using answers to polar questions alone in other contexts is not trivial. Answers are contextualized, and presume that the interrogative question clause and any shared knowledge between the asker and answerer are provided. 

We address the problem of \emph{controllable} rewriting of answers to polar questions into \emph{decontextualized} and \emph{succinct} factual statements. We propose a Transformer sequence to sequence model that utilizes \emph{soft-constraints} to ensure \emph{controllable rewriting}, such that the output statement is semantically equivalent to its \pqa input. 
Evaluation on three separate \pqa datasets as measured through automated and human evaluation metrics show that our proposed approach achieves the best performance when compared to existing baselines.

\end{abstract}

\section{Introduction}\label{sec:introduction}

Polar or Yes/No questions~\cite{huddleston1994contrast} represent one of the main question types, where the answers can be binary, confirming the interrogative clause in the question, with the possibility of containing embedded clauses that may precondition the proposition in the question, or answers can be implicit altogether~\cite{louis-etal-2020-id}.
The examples below show some manifestations of polar questions and answers.

\begin{figure}[h!]
    \centering
    \includegraphics[width=.5\columnwidth]{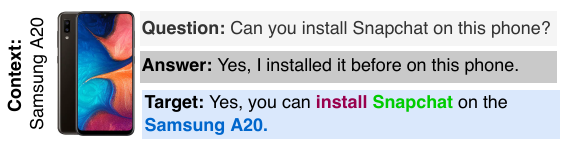}\vspace{-10pt}
    \caption{\small{Example of an input \pqa and the desired rewritten answer into a succinct factual statement.}}
    \label{fig:example}
\end{figure}

\begin{tcolorbox}[left=2pt, right=2pt, colback=gray!5,bottom=2pt,top=2pt]
\scriptsize
\begin{description}
\item{\textbf{Question:}} Did Sandy want coffee?

\item{\textbf{Polar Answer:}} Yes/No.
\item{\textbf{Polar Answer with embedded clauses:}} 
\begin{itemize}[leftmargin=-.2em]
    \itemsep0em
    \item No, Sandy [wants tea]$_{\text{alt.}}$
    \item Yes, Sandy wants coffee, [only if there is cake too]$_{\text{cond.}}$
\end{itemize}
\item{\textbf{Implicit Answer:}} She'd rather have water.
\end{description}
\end{tcolorbox}
On the Web, polar question and answers (\pqa) are present in forums,\footnote{\url{https://www.reddit.com/r/YayorNay/}} e-commerce pages~\cite{Rozen2021}, and on other search related applications~\cite{DBLP:conf/naacl/ClarkLCK0T19,kwiatkowski-etal-2019-natural}. This human curated knowledge remains largely untapped, mainly due to the fact that answers are highly \emph{contextualized} w.r.t their questions, and often are framed in a \emph{personalized} language style. Using such text to answer similar questions, or use them for other applications such as voice-assistants, remains challenging. \revision{Figure~\ref{fig:example} shows an example of a PQA along with a \emph{target} decontextualized answer. The original answer alone is highly ambiguous, and it is not clear what its subject is. }

While most research has focused in answering polar questions~\cite{DBLP:conf/naacl/ClarkLCK0T19,DBLP:journals/corr/abs-2112-07772}, highlighting difficulties of the answering part, no work is done in decontextualizing answers to polar questions. 
In this work, we propose the task of rewriting answers to polar questions (\pqar), by \emph{decontextualizing} them and \emph{rewriting} them into \emph{succinct} factual statements (cf. Figure~\ref{fig:example}), which allows us to leverage the knowledge in \pqa data to answer \emph{similar} questions or questions of \emph{different shapes} such as \emph{wh-*}, where the expected answers are more varied,\footnote{\emph{``What can you install on Samsung A20?''}, can be answered by enumerating through all the decontextualized statements that mention possible applications that can be \emph{installed}.} and furthermore given that they are succinct, answers are interpretable out of their original context and can be used for other downstream applications~\cite{DBLP:journals/tacl/ChoiPLKDC21}.

We propose a \emph{controllable rewriting} approach (\smf), which for an input \pqa generates a decontextualized statement framed in a factual language style using 2nd person narrative. \smf introduces a novel \emph{soft-constraints} mechanism that allows us to achieve controllable rewriting, where for a given set of automatically extracted \emph{constraints} from its input, it ensures constraints satisfaction in the generated statement.

We manually create a dataset of 1500 $\langle$\pqa, \emph{factual statement}$\rangle$ pairs used for training and evaluating different models for the newly introduced task of \pqar. The data is focused on the e-commerce domain~\cite{Rozen2021}, covering a wide range of question domains that Amazon customers ask about different products. 
Our contributions are:
\begin{itemize}
    \item a novel task of \pqar rewriting into factual statements and a novel approach for controllable rewriting through soft-constraints with automatically extracted constraints from input \pqa \revision{based on QA constituency parse trees}; 
    \item a new \pqar dataset for training and evaluating models.
\end{itemize}

\section{Related Work}\label{sec:related_work}

\paragraph{Constrained Text Generation.}
Related work in this domain considers mainly the case of sequence to sequence models~\cite{sutskever2014sequence} and of decoder only based models such as \gpt~\cite{radford2019language}, where the generated output needs to satisfy a given set of \emph{manually} provided constraints.
While pre-trained models such as \tb~\cite{raffel2020exploring} and BART~\cite{lewis2020bart} can be fine-tuned to implicitly capture the co-occurrence between the input and output sequences, they cannot explicitly enforce constraint satisfaction.

To overcome such limitations, controllable rewriting focuses on two types of constraints: \emph{lexical} or \emph{hard} constraints, which consist of a single or sequence of words, enforced on the output. \cite{anderson2017guided} propose Constrained Beam Search (\cbs), which allows only hypotheses that satisfy  constraints. We consider \cbs as our competitor, and show that limiting hypotheses has undesired effects in terms of text quality, and further show limitations of \cbs, where only single token constraints can be efficiently considered.

Similarly, \cite{hokamp2017lexically,juraska2018deep,balakrishnan2019constrained} adapt the inference of seq2seq models to ensure constraint satisfaction. To increase inference efficiency and improve text quality, \cite{wang2021mention} propose Mention Flags (\mf), which trace whether lexical constraints are satisfied. Constraints are explicitly encoded through a \mf matrix, which is added into decoder. When a constraint is satisfied its state in the matrix is changed to \examplenew{satisfied}, thus, providing the model with an explicit signal about constraint satisfaction. Our work is based upon \mf, however, with two significant differences: (1) we propose a mechanism to \emph{automatically} extract constraints, a key component in controllable rewriting, avoiding manual constraint encoding, and (2) we modify \mf s.t \emph{phrases} can be encoded as constraints, and drop the requirements of \emph{one to one} mapping between constraints and the decoded output. We propose \emph{soft mention flags} \smf, where constraint satisfaction is asserted at the semantic level. \revision{Furthermore, through a moving window over the decoded output, we allow our approach to match multi-token constraints to the decoded output. This is another novelty w.r.t MF, where for multi-token constraints all its tokens need to be mapped in the decoded output, thus, allowing our approach to account for paraphrasing.}

\cite{kumar2021controlled} formulate the decoding process as an optimization problem that allows for multiple attributes to be incorporated as differentiable constraints. \cite{qin2022cold} propose Constrained Decoding with Langevin Dynamics (\cold), which treats text generation as sampling from an energy function. We compare and show that our approach outperforms \cold.

\vspace{-10pt}
\paragraph{Yes/No QA.}
\revision{Most works on Yes/No questions are on answering such as QuAC~\cite{choi2018quac}, HotpotQA~\cite{yang2018hotpotqa}, CoQA~\cite{reddy2019coqa}. 
There are several datasets on Yes/No questions ~\cite{clark2019boolq,elgohary-etal-2019-unpack,Rozen2021} that are used for QA.} \cite{Rozen2021} focus on answering product-related questions and construct the Amazon-PQA dataset with a large subset of Yes/No questions. 
Our work is complementary, by decontextualizing answers, it allows for answers to be used on other types of questions (e.g. \emph{wh-*} questions), and additionally since answers are succinct they can be indexed and used in diverse scenarios~\cite{DBLP:journals/tacl/ChoiPLKDC21}.

\vspace{-15pt}
\revision{\paragraph{Question Rewriting.}  Question rewriting in conversational QA~\cite{10.1145/3437963.3441748,DBLP:journals/corr/abs-2210-15777} rewrites questions by resolving co-references from the conversational context. Such works are not comparable to controllable text generation for two main reasons. First, they do not ensure controllability of the generated text. Second, the context is mainly used to augment a given question in a conversational turn, without changing its framing and syntactic shape, as is the case in our work.}

\section{Task Definition and Requirements}
\label{sec:task_definition_pqa_rewriting}

We define the \pqar task of controllable \pqa rewriting into factual statements. 
For an input polar question that is represented by a sequence of tokens $\mathbf{q}=[q_1,\ldots, q_n]$, its answer $\mathbf{a}=[a_1, \ldots, a_n]$, and some context $\mathbf{c}=[c_1,\ldots, c_n]$. The context $\mathbf{c}$ can vary and depending on the domain of PQA data (e.g. it can represent conversation history or in our case some entity title). This input is concatenated into $\mathbf{x}=[\mathbf{q}; \langle\text{\texttt{SEP}}\rangle; \mathbf{a}; \langle\text{\texttt{SEP}}\rangle; \mathbf{c}]$, which is fed to a rewrite function $\mathcal{F}$ that outputs the target statement $\mathbf{y}=[y_1,\ldots, y_n]$, namely $\mathcal{F}(\mathbf{x}) \rightarrow \mathbf{y}$.

\subsection{PQA Syntactic Rewriting Space}\label{subsec:pqa_syntactic_space}
\vspace{-20pt}

\begin{table*}[ht!]
\begin{center}
\resizebox{1\linewidth}{!}{%
\begin{tabular}{p{2cm} p{11cm} p{7cm}}
\toprule
Category & Definition & Examples \\
\hline
\textbf{Explanation} &  The most fundamental shape and basis of all other answers, it consists of the particle \example{yes} or \example{no}, either affirming or negating the preposition in the question. This represent the elliptical type of answer~\cite{kramer2009polarity}. Other cases of \emph{explanation} are answers, where further \emph{evidence} is provided, either by repeating the interrogative clause (in case where there is agreement between the question and answer\footnote{Questions can presuppose either positive or negative answers, hence, an agreement between question and answer refers to their respective polarity.}), or in case of negation, an explanation not present in the question is provided. & 
Yes, \textit{you can install snapchat on the Samsung Galaxy A20 phone.} 
\newline
No, \textit{you cannot install snapchat on the Samsung Galaxy A20 phone.}\\
\\

\textbf{Complement} & In addition to the main response, some answers may include an additional embedded clause that provides further \emph{related} aspect that may interest the asker. & 

Yes, you can install snapchat on the Samsung Galaxy A20 phone. \textit{Also, you can get twitter on it.}  
\newline
No, you cannot install snapchat on the Samsung Galaxy A20 phone. \textit{Also, you can't get twitter on it.} 
\\
\\

\textbf{Condition} & The answer contains an embedded conditional clause~\cite{iatridou1994conditional,williamson2019conditional}, which conditions the truthfulness of either the affirmation or negation of the interrogative clause. Such answers appear when the question is not specific enough. &
Yes, you can install snapchat on the Samsung Galaxy A20 \textit{if it is a smart phone} 
\newline
No, you cannot install snapchat on the Samsung Galaxy A20 \textit{if it is not a smart phone} \\

\\
\textbf{Alternative} & Similar to \emph{explanation}, with the difference that the polarity of the answer is negative w.r.t the question, however, an alternative affirmative proposition is suggested by the answerer. & No, you cannot install snapchat on the Samsung Galaxy A20 phone. \textit{But you can get twitter on it instead.} \\
\bottomrule
\end{tabular}
}
\caption{\small{\pqar syntactic rewriting categories along with their definitions. Example statements for both polarities for the question \emph{``Can you get snapchat on this phone?''} are provided. The highlighted text in the examples shows the embedded clauses for the individual categories.}}
\label{tab:qa_categories}
\end{center}
\end{table*}
\vspace{-30pt}

We now describe the syntactic rewriting space of polar question and answers into factual statements.
The main clause of a polar question is an \emph{interrogative clause}, following specific grammar rules~\cite{keselj2009speech}, e.g., $q \rightarrow $ \texttt{AUX NP VP} (among many other context free grammar rules).%

Answers to polar questions can take several different syntactic shapes~\cite{kramer2009polarity,polar_answers,HOLMBERG201331}.
Furthermore, as there is an asynchronous relation between the asker and answerer in online settings, there is often a lack of conversational context, resulting in more elaborate answers.
We devise a taxonomy of answer types, which correspondingly determine also the shape of the rewritten statement. Table~\ref{tab:qa_categories} defines the different categories along with example generated statements for each category.

The cases in Table~\ref{tab:qa_categories} represent frequent syntactical manifestations of PQAs on the Web, and we will use them as guidance for dataset collection (cf. Section \ref{sec:dataset_collection}) and evaluation (cf. Section \ref{subsec:baseline}).

\section{Approach}

\begin{figure*}[!t]
    \centering
    \includegraphics[width=1.\linewidth]{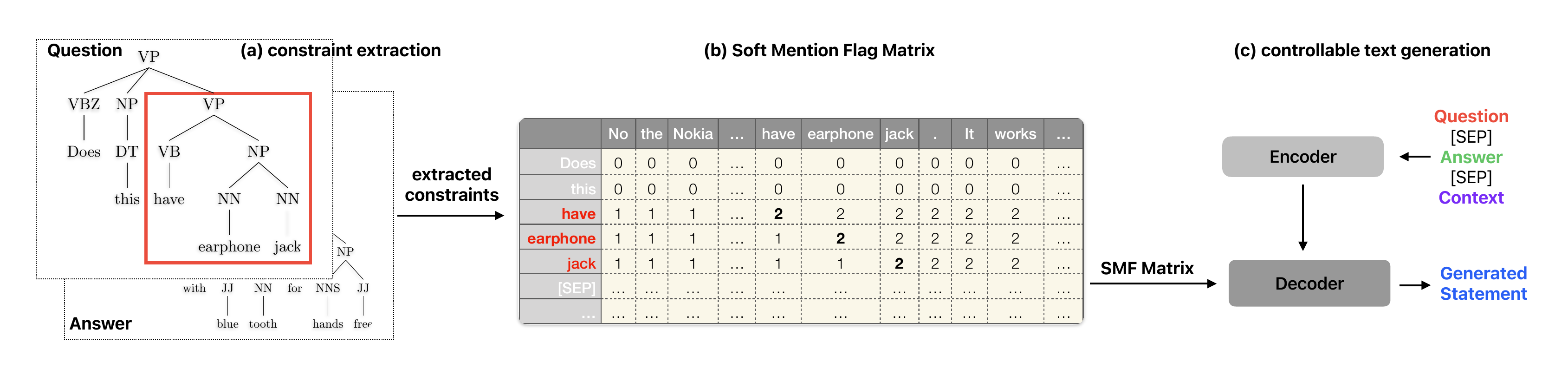}\vspace{-10pt}
    \caption{\small{Overview of our proposed controllable rewriting approach. In (a) we extract automatically constraints from constituency parse trees, then in (b), the constraints together with the input text that goes to \emph{encoder} is encoded in the soft-mention flag matrix, which is then provided as input in (c) to the decoder for controllable generation.}}
    \label{fig:pipeline}
\end{figure*}

Figure~\ref{fig:pipeline} shows an overview of our approach, a sequence to sequence (seq2seq) model based on the T5 transformer model~\cite{radford2019language}. It has two main components that \emph{ensure controllable rewriting}: (i) automated constraint extraction, and (ii) controllable rewriting through soft-constraints.

\subsection{Automated Constraint Extraction}\label{subsec:constraint_extraction}
Unlike in MF~\cite{wang2021mention}, where constraints are provided manually, we propose an automated constraint extraction approach based on constituency parses of the input PQA.

\revision{Our goal in controllable rewriting is for the target statement to contain the interrogative clause from the question, and the corresponding affirmation/negation clauses in the answer, along with the any embedded conditional or alternative clauses. This boils down to two main tasks for constraint extraction: (1) determining the input tokens that \emph{must be present} in the output, and (2) ensuring that the decoder satisfies such constraints.}

Following the syntactic rewriting cases in Table~\ref{tab:qa_categories}, we extract constraints, shown in order of importance. (the algorithm is provided in the paper's appendix).
\begin{description}[leftmargin=*]
\item {\textbf{Noun Phrases (NP)}:} NPs identify the subject of the question. In the answer they help us identify the matching clause with the assigned \emph{polarity} and \emph{explanation} to the question's preposition. NPs are used as constraints only if they are not embedded as children of non-NP constituents. 
\item {\textbf{Verb Phrases (VP)}:} VPs on the other hand identify the information need as defined by the verb serving as the root of the constituent.
\item {\textbf{Other Phrases}:} The rest of the extracted phrases as constraints are prepositional phrases (PP), adverbial phrases (ADVP), and adjective phrases (ADJP). PP and ADJP  provide further details about the  information in an NP, whereas ADVP provide further information about the verb in the question's interrogative clause and assert its polarity.

\end{description}

\subsection{Soft Mention Flags}\label{subsec:encoding_constraints}

To assess if the extracted constraints are satisfied in the generated output, we must account for two factors: (i) input constraints may be expressed differently in the output (\emph{paraphrases}, \emph{synonyms} etc.), and (ii) there is no one-to-one mapping between input constraints and the generated statement.

We enforce our model to \emph{satisfy} the extracted constraints in \S\ref{subsec:constraint_extraction} by constructing a soft-mention flags matrix $\mathbf{M}_{k\times l} \in \{0, 1, 2\}$, where rows represent input \pqa tokens and columns are the output tokens. For each input token $x_i$, $\mathbf{M}$ holds a value between $\{0, 1, 2\}$, where 0 is for tokens not part of any constraint, 1 for token part of a constraint but not satisfied, and 2 for tokens part of a constraint and satisfied in the output $\mathbf{y}$. 
{\begin{equation}\label{eq:sf_matrix}\small
    M_{\mathbf{x}_{i}, \mathbf{y}_{:t}}=
        \begin{cases}
          0 & x_{i} \text{ is not part of a constraint}\\
          1 & x_{i} \text{ is not mentioned in } \mathbf{y}_{:t}\\
          2 & x_{i} \text{ is mentioned in } \mathbf{y}_{:t}
        \end{cases}
\end{equation}}
\revision{As in \cite{wang2021mention}, we inject the soft mention flag matrix, $\mathbf{M}$, in the model's decoder layers. $\mathbf{M}$ is represented through two embedding types: 1) key embeddings, $\mathbf{M}^{k}=E_{k}(M)$ and value $\mathbf{M}^{v}=E_{v}(M)$ where $E_{k}$ and $E_{v} \in \mathbb{R}^{3\times dim}$. These embeddings are injected between the encoder output $\mathbf{h}^{e}$ and the decoder input $\mathbf{h}_{t}^{d}$ in the cross multi-head attention module (cf. Eq~\ref{eq:ca}).}
{\begin{equation}\label{eq:ca}\small
\begin{split}
\text{\texttt{CA}}&(\mathbf{h}_{t}^{d}, \mathbf{h}^{e}, \mathbf{M}^{k}, \mathbf{M}^{v}) = \\ &F(W_{q}h_{t}^{d}, W_{k}h^{e}, W_{v}h^{e}, \mathbf{M}^{k}, \mathbf{M}^{v})
\end{split} 
\end{equation}}
where $F$ is a self-attention function with soft mention flag embeddings defined below.
{\begin{align}
\small{F(q, k, v, \mathbf{M}^{k}, \mathbf{M}^{v})_{j} = \sum_{i=1}^{l_{x}}\alpha_{i,j}(v_{i}+\mathbf{M}_{i,j}^{v})} \\
\small{\alpha_{i,j} = \text{\texttt{softmax}}\left(\frac{q_{i}(k_{i}+\mathbf{M}_{i,j}^{k})^{T}}{\sqrt{dim}}\right)}
\end{align}}

$\mathbf{M}$ provides the seq2seq model with \emph{explicit} signal about the decoded tokens.  Whenever input tokens are marked with 0, the model performs standard conditional decoding. Otherwise, if a token is part of a constraint (set to 1 in $\mathbf{M}$), it represents an explicit signal to decode sequences such that the corresponding values in $\mathbf{M}$ are changed to 2. 

\revision{With the explicit means to signal \emph{what} part of the input composes a constraint that needs to be met in the decoded output, next, we describe how we establish if a constraint is satisfied, while taking into account that it can undergo \emph{syntactic} and \emph{lexical} changes in the decoded output. We propose two strategies to encode constraint satisfaction.}

\paragraph{\textbf{Semantic Constraint Satisfaction:}}

For an input constraint all its tokens $\mathbf{c}_i=\{x_m,\ldots, x_n\}$\footnote{A sequence of one or more consecutive tokens as extracted from the constituency parse tree} are initialized with 1 in $\mathbf{M}$ (i.e.,  constraints not been satisfied). 
At each generation step $t$, we assess whether the constraint $\mathbf{c}_i$ is satisfied. To ensure constraint satisfaction accuracy, we consider only tokens within a specific \emph{window length} (with window size equal to $|\mathbf{c}_i|$) of preceding tokens in the output $\mathbf{y}_{k,l}$\footnote{Where $\forall\;\; 0 \leq k \leq t-|\mathbf{c}_i|$ and $k < l \leq t$}. A constraint is satisfied if the semantic similarity between \texttt{sim}$(\mathbf{y}_{k,l}, \mathbf{c}_i)$, computed as the cosine similarity between the sentence representations~\cite{DBLP:conf/emnlp/ReimersG19} of $\mathbf{y}_{k,l}$ and $\mathbf{c}_i$ meets two specific thresholds: (1) threshold $a$, where $\text{\texttt{sim}}_t > a$, (2) threshold $b$ of the difference between the current similarity score and the score in step $t-1$, namely $\text{\texttt{sim}}_t - \text{\texttt{sim}}_{t-1} > b$. Once $\mathbf{y}_{:t}$ meets both thresholds, all tokens of $\mathbf{c}_i$ are changed to $2$ in $\mathbf{M}$.

\paragraph{\textbf{Example:}} Table~\ref{tab:smf_example_1_2} (a) shows an example on the initialization and the updates on the soft mention flag matrix $\mathbf{M}$. For an input sequence $\mathbf{x}=[$The, screen, has, full, touchscreen, function$]$, $\mathbf{c}=$ \emph{``has full touchscreen function''}, is a constraint that needs to be satisfied in the output. The flags for $\mathbf{c}$ at step 0 are initialized with 1, $\mathbf{M}(\mathbf{c}, y_{:0})=[0,0,1,1,1,1]$, given that the output is $\langle \text{\texttt{SEP}}\rangle$. The top row shows the  similarity score for each step. At step $6$,  $\mathbf{M}(\mathbf{c}, y_{:0})=[0,0,2,2,2,2]$ since the constraint has been covered in the current output sequence.
\begin{table}[!htb]
    \begin{subtable}{.5\linewidth}
      \centering
      \resizebox{0.95\textwidth}{!}{
        \begin{tabular}{c cccccccc}
        \toprule
        \makecell[c]{\texttt{sim}} & 0 & 0.21 & 0.26 & 0.28 & 0.26 & 0.37 & \textbf{0.85} & 0.76\\
         \hline
         & $\langle$ \texttt{SEP} $\rangle$ & Dell & Laptop & \underline{comes} & \underline{with} & \underline{full} & \underline{touchscreen} & .\\
        \midrule
        The & 0 & 0 & 0 & 0 & 0 & 0 & 0 & 0\\
        screen & 0 & 0 & 0 & 0 & 0 & 0 & 0 & 0\\
        \textbf{has} & 1 & 1 & 1 & 1 & 1 & 1 & \textbf{2} & 2\\
        \textbf{full} & 1 & 1 & 1 & 1 & 1 & 1 & \textbf{2} & 2\\
        \textbf{touchscreen} & 1 & 1 & 1 & 1 & 1 & 1 & \textbf{2} & 2\\
        \textbf{function} & 1 & 1 & 1 & 1 & 1 & 1 & \textbf{2} & 2\\
        \bottomrule
        
        \end{tabular}}
        \caption{}
    \end{subtable}%
    \begin{subtable}{.5\linewidth}
    
              \centering
              \resizebox{0.95\textwidth}{!}{
                \begin{tabular}{c ccccccccccc}
        \toprule
        
         & $\langle$ \texttt{SEP}$\rangle$ & Dell & XPS & can & be & shipped & by & \underline{us} & to & Brazil & .\\
        \midrule
        \textbf{We} & 2 & 2 & 2 & 2 & 2 & 2 & 2 & 1 & 1 & 1 & 1\\
        can & - & - & - & - & - & - & - & - & - & - & -\\
        ship & - & - & - & - & - & - & - & - & - & - & -\\
        to & - & - & - & - & - & - & - & - & - & - & -\\
        Brazil & - & - & - & - & - & - & - & - & - & - & -\\[1.45ex]
        
        \bottomrule

    \end{tabular}}
    \caption{}
    \end{subtable}\vspace{-10pt}
    \caption{\small{ (a) Constraints are marked in bold. Underlined tokens are included in the sliding window that assesses constraint satisfaction. (b) 1st person pronouns are initialized with 2 then changed to 1, when a 2nd person pronoun word is generated.}}
    \label{tab:smf_example_1_2}
\end{table}
\vspace{-20pt}

\paragraph{Factual Style Constraints:} \revision{To frame the output statement in a factual style, first person narratives are transformed into a second person narrative. Such a seemingly small change (i.e., \emph{1st} to \emph{2nd} person pronouns),  incurs a series of syntactic rewrite operations required to ensure coherence of the output statement. Table~\ref{tab:smf_example_1_2} (b) provides an example. In this case, we represent in the $\mathbf{M}$ 1st person pronouns with score 2, and convert them to 1, once the model has generated a second person pronoun in the output. The reason for reverting the order from \emph{satisfied} constraint to \emph{not satisfied} is to avoid any confusing behavior between the satisfaction of input constraint extracted through the automated constraint extraction and framing constraints.}

\section{\pqar Data Collection}\label{sec:dataset_collection}

We now describe our data collection process, which is based on the Amazon Product Question Answers dataset~\cite{Rozen2021}, which contains a diverse set of product-specific PQAs generated by customers. The dataset contains 10M questions about 1.5M products, from which we focus only on yes/no questions. Each instance consists of the question text $q$, answer text $a$, and product name $c$, which represents the context in our case.

\subsection{PQA Rewriting through Crowdsourcing}\label{subsec:crowdsourcing}

We recruited 10 expert annotators to collect ground truth for 1,500 instances. The data is uniformly distributed across the categories in Table~\ref{tab:qa_categories} and covers 11 domains.

The task is designed to be decomposable into stages to ensure annotation reliability. Annotators follow a series of guidelines (cf.Table~\ref{tab:crowdsource_example}), allowing them to first map the input \pqa into one of the pre-defined categories, after which they determine the answer polarity (step \textbf{S1}), then in \textbf{S2} the output category is determined (e.g. \emph{Alternative}), in \textbf{S3} the anaphoric expressions are replaced with the context information, and finally in \textbf{S4} the statement is framed in 2nd person narrative. 
\begin{table*}[h!]
\begin{center}
\resizebox{1\linewidth}{!}{
\begin{tabular}{c|l}
\toprule
\textbf{Q} & Does this monitor have a camera?\\
\textbf{A} & No. But it is a wide screen. I've been connecting a game console to it.\\
\textbf{C} & Dell 27-inch Full HD 1920$\times$1080 Widescreen LED Professional Monitor \\
\midrule
\textbf{S1} & \textbf{\textit{No}}.\\
\textbf{S2} & No, \textbf{\textit{this monitor doesn't have a camera. But it has a wide screen and I've been connecting a game console to it}}.\\
\textbf{S3} & No, \textbf{\textit{the Dell 27-inch Full HD monitor}} doesn't have a camera. But it has a wide screen and I've been connecting a game console to it.\\
\textbf{S4} & No, the Dell 27-inch Full HD monitor doesn't have a camera. But it has a wide screen and \textbf{\textit{you can connect it with a game console}}.\\

\bottomrule

\end{tabular}
}
\caption{\small{An example to illustrate how annotators perform the Yes/No QA rewriting task. }}
\label{tab:crowdsource_example}

\end{center}
\end{table*}

\vspace{-40pt}\section{Experimental Setup}\label{sec:experiment}

\subsection{Datasets}
\label{data}

\noindent\textbf{\pqar:} We randomly split the \pqar dataset into 1000/100/400 for train/dev/test respectively. \pqar is used for training and evaluation.

\noindent\textbf{Reddit:} We leverage a sub-forum of Reddit of polar questions\footnote{{\url{https://www.reddit.com/r/YayorNay/}}} and on a randomly sample of 50 QA pairs assess zero-shot generalization performance. 

\noindent\textbf{SemEval:} From SemEval-2015 Task 3\footnote{{\url{https://alt.qcri.org/semeval2015/task3/index.php}}} we randomly sample 50 pairs of Yes/No question and answers for zero-shot evaluation.

\subsection{Baselines}\label{subsec:baseline}
\revision{Approaches are trained on the \pqar dataset in \S\ref{data}.}

\noindent\textbf{T5}~\cite{raffel2020exploring}. We use the T5 model as our baseline, which in turn serves as an ablation of \smf without the soft-mention flags module.

\noindent\textbf{CBS}~\cite{anderson2017guided}. We use \tb with constrained beam search during the decoding phase. The constraints represent phrases extracted in \S\ref{subsec:constraint_extraction}.

\noindent\textbf{GPT-2}~\cite{radford2019language}. We adopt \gpt and fine-tune it on \pqar training set.

\noindent\textbf{COLD}~\cite{qin2022cold}. A \gpt based decoding method by sampling from an energy function.

\noindent\textbf{MF}~\cite{wang2021mention}. Originally proposed the addition of mention flags into sequence to sequence decoders, and focuses only on lexical constraints. The constraints in this case, contrary to the original paper that are provided as input, here the constraints are extracted automatically (cf. \S\ref{subsec:constraint_extraction}).

\noindent\textbf{Our Approach -- \smf:} We distinguish two models of our approach: (1) \textbf{\smf} where for controllability we rely only on the extracted constraints in \S\ref{subsec:constraint_extraction}, and (2) \textbf{\smfs}, where in addition to the extracted constraint, we additionally encode the target style constraints (\S\ref{subsec:encoding_constraints}).

\subsection{Evaluation Metrics}\label{subsec:metrics}

\paragraph{\textbf{Automated Metrics:}} To assess the closeness of the generated statements with respect to the ground-truth statements generated by human annotators, we use BLEU \cite{papineni-etal-2002-bleu}, ROUGE \cite{lin-2004-rouge} and F1-BertScore \cite{Zhang*2020BERTScore:}.

\paragraph{\textbf{Human Evaluation:}} Automated metrics cannot capture the nuanced aspects and natural variation of the task. We design two human evaluations, where annotators assess 50 randomly sampled input-output pairs for on:

\textbf{Statement Syntactic Clause Coverage:} we assess if the statement, depending on its category (cf. Table~\ref{tab:qa_categories}) has coverage of the embedded clauses.

\textbf{Statement Correctness and Coherence:} we assess if a statement: (i) contains the correct polarity, (ii) mentions the input context, (iii) framing is in 2nd person narrative, (iv) does not contain information not present in the input PQA; (v) is grammatically correct and coherent, and (vi) is equivalent to its ground-truth counterpart.

\section{Experimental Results}\label{sec:rq3}

\paragraph{Overall Performance.} Table~\ref{tab:main_results} (a) shows the evaluation results for all models on the \pqar test set. We observe that our proposed \smf approach achieves the highest performance across all evaluation metrics. A negligible difference is noted between \smf and \smfs. However, as we show in \S\ref{subsec:human_evaluation} the \smfs attains more coherent statements. 

From the baselines, the closest to our approach is \mf. However, note here that we adapt \mf using our extracted constraints, contrary to \cite{wang2021mention} where the constraints are provided manually, an approach that does not scale. 

\revision{Alternatively, if we do not use our proposed automated constraint extraction approach, one could  provided as constraints to \mf tokens that overlap between the question and answer. Using such constraints causes the performance of \mf to drop by 27\% in terms of BLEU. This highlights that \emph{what} composes as input constraints is key, and noisy constraints can yield results worse than seq2seq models without constraints. }

\cbs reveals that our task does not simply involve copying tokens, but rather involves a series of syntactic and semantic transformations of the input. The difference between \cbs and \smf is 7.4\% in terms of BLEU and ROUGE-L scores. This shows that enforcing the decoder to output certain tokens has a negative effect, as \cbs performs worse than \tb. This is another indication that our approach, where the constraint satisfiability is done at the semantic level is appropriate. \revision{Equivalent clauses from the input can be transformed and in cases question and answer clauses can be combined into a single clause, hence, causing difficulties to establish such mappings.}

\cold and \gpt achieve poor performance, highlighting the need for larger amount of training to learn the task.

\begin{table}[!htb]
    \begin{subtable}{.5\linewidth}
      \centering
      \resizebox{0.8\textwidth}{!}{
       \begin{tabular}{l ccc}
        \toprule
        \textbf{} & \textbf{BLEU} & \textbf{ROUGE-L} & \textbf{BertScore} \\
        \midrule
        \gpt & 32.8 & 55.0 & 93.2 \\
        \cold & 28.1 &  51.3 & 92.0 \\
        \tb & 50.6 & 67.9 & 95.3 \\
        \cbs & 45.2 & 62.7 & 94.2 \\
        \mf & 51.2 & 68.5 & 95.4 \\
        \midrule
        \smf & \textbf{52.6} & \textbf{69.5} & \textbf{95.5} \\
        \smf-Style & 52.5 & 68.9 & 95.4 \\
        \bottomrule
        \end{tabular}
        }
        \caption{}
    \end{subtable}%
    \begin{subtable}{.5\linewidth}
    
              \centering
              \resizebox{0.95\textwidth}{!}{
             \begin{tabular}{ll ccc}
\toprule
& & \textbf{ BLEU } & \textbf{ ROUGE-L } & \textbf{ BertScore} \\
\midrule
\multirow{3}{*}{complement} & \mf & 40.3/43.3 & 68.0/70.1 & 95.2/95.5 \\
& \smf & 45.0/45.8 & 71.3/71.5 & 95.6/95.6 \\
& \smfs & 45.6/46.8 & 71.1/73.8 & 95.9/96.0 \\
\bottomrule

\multirow{3}{*}{condition} & \mf & 35.3/36.7 & 62.0/62.8 & 94.7/94.7 \\
& \smf & 34.5/39.5 & 61.4/63.9 & 94.7/95.0 \\
& \smfs & 36.5/37.9 & 62.3/64.1 & 95.0/95.1 \\
\bottomrule

\multirow{3}{*}{alternative} & \mf & 40.8/41.1 & 69.5/69.9 & 96.1/96.3 \\
& \smf & 44.1/44.8 & 70.0/71.0 & 96.0/96.1 \\
& \smfs & 43.5/44.5 & 69.2/71.5 & 96.3/96.4 \\
\bottomrule
\end{tabular}
}
    \caption{}
    \end{subtable} \vspace{-20pt}
    \caption{\small{ (a) {Overall performance of the different models on the \pqar test set. \revision{\mf, \smf and \smf-Style obtain significantly better results (p$<0.01$ as per t-test) than the rest of the competitors.}}. (b) Impact of input constraints on output quality. The first score represents the case without the extracted input constraints, while the second score represents the case with constraints. }}
    \label{tab:main_results}
\end{table}

\begin{figure}[h!]
    \centering
    \includegraphics[width=0.7\columnwidth]{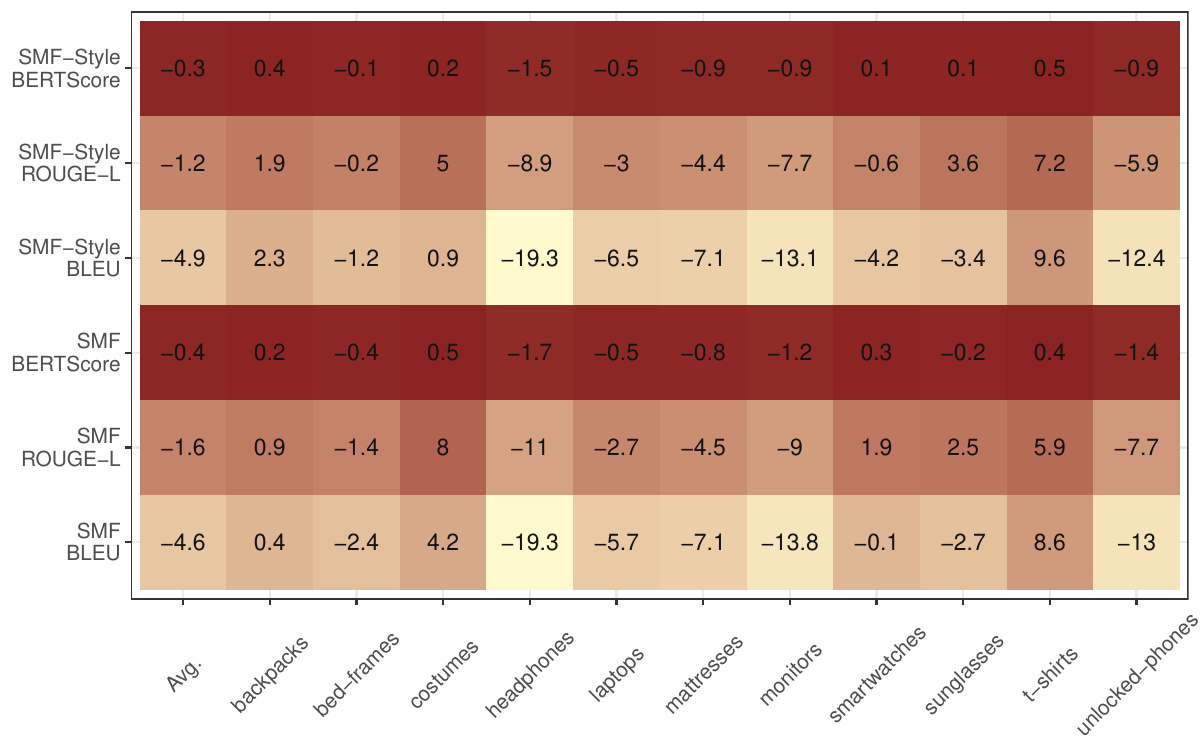}\vspace{-10pt}
    \caption{\small{Out-of-domain performance \revision{ on the test set of the \pqar dataset on  different domains.}}}

    \label{tab:out_of_domain}
\end{figure}

\paragraph{\textbf{Out-of-Domain Performance.}} Figure \ref{tab:out_of_domain} shows the out-of-domain performance of \smf and \smfs. For the 11 domains in the \pqar dataset, we consider a leave-one-out domain evaluation, showing the zero-shot performance.  The results show that: (1) both approaches generalize well on out-of-domain data, with most domains obtaining comparable performance to the in-domain performance. (2) comparing \smf and \smfs, we note that \smfs achieves better performance on statement framing style, implying the effectiveness of style constraints.

\paragraph{\textbf{Input Constraint Coverage Impact.}} Table~\ref{tab:main_results} (b) shows the impact of constraints, \revision{namely the presence of constraints as input} on the generated output. The results show the impact on different \pqar categories, for approaches that make use of the input constraints.\vspace{-15pt}

\vspace{-20pt}

\subsection{Human Evaluation}\label{subsec:human_evaluation}
To complement the automatic metrics, we report the results of our two human studies \revision{carried out by two human annotators}. 

\paragraph{\textbf{Statement Syntactic Clause Coverage:}}

Table~\ref{tab:human_eval_statement_1_2} (b) shows the results on the evaluation of statements containing the different clauses according to their \pqa syntactic shape. Namely, if an input belongs to the \emph{condition} syntactic shape, the generated statement should contain the \emph{explanation} and the embedded \emph{condition} clause. 

Except for \cbs and \cold, all approaches do fairly well in incorporating the \emph{explanation} clause in their output. This is intuitive as this is present in all input PQAs, and contains information that is present in the question and often in the answer as well.  

One of the key embedded clauses, the \emph{condition} clause, represent one of the most challenging scenarios to be decoded into the target statement. This clause has a pivotal role in conditioning the polarity of the answer, and thus, failure to decode this in the statement can lead to erroneous answers provided to a question. \smf approaches achieve the highest scores, with \tb having a 10\% point difference. 
One notable case is that of \emph{alternative} clause, where the baseline \tb achieves perfect coverage, with \mf and \smf following with 92.9\% coverage. 

Overall, the best coverage is achieved by \smf, with 85\% coverage across all types. This shows the importance of extracted input constraints based on our approach in \S\ref{subsec:constraint_extraction}, where the models \mf, \smf, and \smfs, achieve highest coverage among all competing approaches.\vspace{-10pt}

\begin{table}[!htb]
    \begin{subtable}{.5\linewidth}
      \centering
      \resizebox{0.8\textwidth}{!}{
       \begin{tabular}{l cccc c}
        \toprule
        \textbf{} & \textbf{ Explanation } & \textbf{ Complement} & \textbf{ Condition} & \textbf{Alternative} & \textbf{Overall} \\
        \midrule
        \gpt &91.0	 & 55.6 &	46.7 &	64.3 &	58.0 \\
        \cold & 65.0	 &	44.4 &	43.3 &	35.7 &	45.0 \\
        \tb & 96.0	 &	75.0 &	70.0 &	\textbf{100.0} &	79.0 \\
        \cbs & 86.0	 &	66.7 &	60.0 &	85.7 &	72.0 \\
        \mf & \textbf{99.0}	 &	\textbf{83.3} &	66.7 &	92.9 &	82.0 \\
        \smf &98.0 & \textbf{83.3}	 &	\textbf{80.0} &	92.9 &	\textbf{85.0} \\
        \smf-Style &	98.0 &	80.6 &	\textbf{80.0} &	78.6 &83.0	 \\
        
        \bottomrule
        \end{tabular}
        }
        \caption{}
    \end{subtable}%
    \begin{subtable}{.5\linewidth}
    
              \centering
              \resizebox{0.95\textwidth}{!}{
                    \begin{tabular}{l rrrrrrr}
        \toprule
        & \textbf{Polarity} & \textbf{Coverage } & \textbf{Style} & \textbf{Relevance} & \textbf{Syntactic} & \textbf{Coherence} & \textbf{Equivalence} \\
        \midrule
        \gpt &	0.93 & 0.94 &	0.98  &	0.67 &	0.84 &	0.69  & 0.36 	 \\
        \cold & 0.80 	 & 0.75 	 &	0.98  &	0.52  & 0.43 	 &	0.42  &	0.19  \\
        \tb & 0.96 	 &\textbf{1.0}	 &	0.97  &	0.94  &	0.95  &	0.91  &	0.74  \\
        \cbs & 0.82 	 &	0.87  &	0.95  &	0.82  &	0.68  &	0.75  &	0.56 \\
        \mf & 0.94 	 &	0.98  &	\textbf{0.99} &	0.94 &	0.98  &	0.91  & 0.69 	 \\
        \smf & 0.95 	 & 0.98 	 &	0.93  &	0.91  &	0.94  &	0.92  &	\textbf{0.77 }\\
        \smfs &	\textbf{0.97} &	0.99  &	0.96  &\textbf{0.96} &	\textbf{0.99} &	\textbf{0.98} &	0.76  \\
        \bottomrule
        \end{tabular}
        }
    \caption{}
    \end{subtable} \vspace{-20pt}
    \caption{\small{ (a) Clause type coverage on the output statement. (b) Correctness and coherence scores of the generated statements by the different approaches. }}
    \label{tab:human_eval_statement_1_2}
\end{table}

\vspace{-30pt}
\paragraph{\textbf{Correctness and Coherence Evaluation:}}

Table~\ref{tab:human_eval_statement_1_2} (a) shows the human evaluation results for the syntactic correctness and coherence of statements, where several detailed aspects are considered. This study highlights the task complexity and provides an overview of which sub-tasks the different models are able to do reliably. 

On \emph{polarity}, most seq2seq based models, such as \tb, \mf, \smf and \smfs,  obtain high polarity accuracy, with \smfs achieving an accuracy of 97\%. Given that answers are contextualized w.r.t the question and the input context, here the product name, we note that similarly, all \tb based models obtain a good coverage of the context, with \tb achieving 100\% coverage. In terms of context coverage, the models are required to chose from the lengthy product names, a succinct phrase that allows the human annotators to identify correctly the product name. 

On \emph{factual style}, \mf obtains the highest accuracy, however, at a cost that not always the generated statements are semantically coherent. When we consider \smfs, which encodes as constraints the framing style, although the accuracy is lower in terms of outputting statements in 2nd person narrative, in terms of coherence, we obtain the highest accuracy. Coherence, represents a more global measure, which has high importance, given that incoherent statements are not suitable to be provided as answers. 

Finally, in terms of equivalence between the generated statement and the ground truth, both \smf and \smfs, obtain the highest scores. This shows that our approaches are able to jointly optimize for the numerous sub-tasks of controllable rewriting.

\subsection{Evaluation on Other Community PQA}
Finally, we evaluate the zero-shot performance of \smf and \smfs on \reddit and \semeval w.r.t the statement correctness and coherence. 

Table~\ref{tab:other_community_qa} shows that our approach can perform well when applying to other data sources under the zero-shot setting. The results of this setting imply that the modeling of our rewriting task is not simply learning the superficial lexical content, but learning the rewriting strategies to restructure the input PQA regardless of domains. This finding makes our rewriting task and approach promising to generalize to other domains without requiring extra annotations. 
\vspace{-10pt}

\begin{table}[ht]
\begin{center}
\resizebox{.4\linewidth}{!}{
\begin{tabular}{l rr rr}
\toprule
\textbf{} & \multicolumn{2}{c}{\reddit} & \multicolumn{2}{c}{\semeval} \\
\textbf{} & \smf & \smf-Style &  \smf & \smf-Style \\
\midrule
Polarity & 1.0 & 1.0 & 0.90 & 0.95 \\
Context Coverage & 1.0 & 1.0 & 1.0 & 1.0 \\
Factual Style & 0.95 & 1.0 & 1.0 & 1.0 \\
Relevance & 1.0 & 1.0 & 0.95 & 1.0 \\
Syntactic Correctness & 0.95 & 1.0 & 1.0 & 1.0 \\
Coherence & 0.90 & 0.95 & 0.95 & 0.95 \\
Equivalence & 0.85 & 0.80 & 0.80 & 0.80 \\

\bottomrule
\end{tabular}
}
\caption{\small{Results on \reddit and \semeval datasets.}}
\label{tab:other_community_qa}
\end{center}
\end{table}

\vspace{-50pt}
\section{Conclusion}\label{sec:conclusion}

We introduced the task of rewriting Yes/No question and answers into succinct decontextualized statements, and defined several desiderata  determining how the input \pqa is reorganized and framed into the factual statement. This task enables us to explore knowledge from  community \pqa, unlocking the highly contextualized answers to answer other question shapes, and furthermore making them retrievable. For this task, with the help of expert annotators, we curated a dataset of 1500 input PQA and the target statements covering 11 domains from Amazon's PQA dataset.

Next, we introduced an approach for controllable rewriting, achieved through automatically extracted constraints from the input, which are encoded into our approach using a \emph{soft mention flag} matrix, allowing us at the semantic level to map constraints to the generated statements.

Finally, empirical evaluations showed that our approach outperforms a series of competitors in both automated and human evaluation metrics.

\section*{\centering Appendix}

\paragraph{\textbf{Constraint Extraction From Parse Trees:}}
Algorithm~\ref{algo1} outlines the steps undertaken to extract constraints from the extracted constituency parse trees from the input \pqa.
\begin{algorithm}[htb]
    \scriptsize
    \caption{Constraint Extraction}
    \label{algo1}
    \begin{algorithmic}[1]
        \REQUIRE Yes/No question $Q$, answer $A$. \\
        \ENSURE ~~\\
        \STATE Extract all NP from the input using constituency parsing tree. \\
        \STATE Exclude PRON from NP in Step 1. \\
        \FOR{each noun phrase}
            \IF{the parent node is labeled as [`VP', `PP', `ADVP', `ADJP']}
                \STATE Add the phrase that belongs to this parent node to the constraint list.
            \ELSE
                \IF{the parent node is labeled as `NP'}
                    \STATE Add the phrase that belongs to the current node to the constraint list.
                \ENDIF
            \ENDIF
        \ENDFOR
        \RETURN constraint list
    \end{algorithmic}
\end{algorithm}\normalsize

\paragraph{\textbf{Constraint Extraction Accuracy:}}
We sample 100 instances in the \pqar dataset which uniformly cover all the different syntactic \pqar categories from Table~\ref{tab:qa_categories}, and ask annotators whether: (1) the extracted constraints from the question or the answer capture the question's intent, and (2) if the extracted constraints cover the different embedded clauses. 
For the  the first part, we see that the extracted constraints from both the question and the answer have a good coverage of 87\% on the question's intent. 

\begin{itemize}[leftmargin=*]
\itemsep0em
    \item \textbf{Q1}: \examplenew{Do the extracted question constraints cover fully the question's intent?} = 87.1\%
    
    \item \textbf{Q2}: \examplenew{Do the extracted answer constraints cover fully the question's intent?} = 87.7\%
\end{itemize}

In the second part, the extracted constraints cover well on explanation and complement. Especially, explanation covers nearly 97\% of the cases. A lower coverage is reported for constraints covering the condition and alternative clauses in the answer. This is mainly due to the fact that these clauses use pronouns, thus, increasing the likelihood of missing those simplified constituents. This represents an important future research direction to have a more robust algorithm with better coverage of all answer's embedded clauses.  More specifically, we obtain the following scores: Explanation=0.97, Complement=0.7, Condition=0.52, and Alternative=0.40.

\paragraph{\textbf{Human Evaluation Analysis:}}
\revision{
Two annotators evaluated 50 randomly chosen Yes/No QA pairs. All judgments are on a single scale (i.e., \emph{binary}). For reliability, each instance is assessed by both annotators. In case of ties, a third annotation was collected. }

\revision{\noindent\textbf{Statement Syntactic Shape:} In this study, the inter-rater agreement rate was 87.5\%. This represents a high agreement rate and shows that annotators agree on what embedded clauses are covered in the generated target statement.}

\revision{\noindent\textbf{Statement Correctness and Coherence:} The inter-rater agreement was measured separately on the seven different questions, with an agreement of \emph{Polarity}=64\%, \emph{Coverage}=64\%, \emph{Style}=65\%, \emph{Relevance}=62.3\%, \emph{Syntactic}=64\%, \emph{Coherence}=62.3\%, and \emph{Equivalence}=52\%, respectively.}

\revision{Overall, the agreement rates are high and similar for most of the questions, with the only exception being \emph{Equivalence}. In this question, the annotators were asked to assess if two statements represent semantically equivalent information. This  yields a lower agreement rate, as annotators may comprehend the statements differently, and additionally the presence or absence of information on one of the statements can cause the annotators to perceive the equivalence differently. 
}

\bibliographystyle{splncs04}
\bibliography{anthology}

\end{document}